\documentclass[runningheads]{llncs}

\usepackage{eccv}

\usepackage{eccvabbrv}

\usepackage{graphicx}
\usepackage{booktabs}

\usepackage[acronym]{glossaries}
\newacronym{ib}{IB}{Information Bottleneck}
\newacronym{dnn}{DNN}{deep neural network}
\newacronym{kl}{KL}{Kullback-Leibler}
\newacronym{dpi}{DPI}{Data Processing Inequality}
\newacronym{fgsm}{FGSM}{Fast Gradient Sign Method}

\newacronym{bim}{BIM}{Basic Iterative Method}

\newacronym{mim}{MIM}{Momentum Iterative Method}

\newacronym{pgd}{PGD}{Projected Gradient Descent}
\newcommand{\pgd}{\gls{pgd}\xspace}
\newacronym{nes}{NES}{Natural Evolutionary Search}

\newacronym{eatk}{EVO}{Evolutionary Attack}

\newacronym{sopt}{S-OPT}{Sign-OPT Attack}

\newacronym{hsja}{HSJA}{Hop-Skip-Jump Attack}

\newacronym{ec}{EC}{Edge Computing}
\newacronym{kd}{KD}{Knowledge Distillation}
\newacronym{sc}{SC}{Supervised Compression}
\newacronym{es}{ES}{Entropic Student}
\newacronym{bf}{BF}{BottleFit}
\newacronym{jc}{JC}{JPEG Compression}
\newacronym{qt}{QT}{Quantization}
\newacronym{sota}{SOTA}{state-of-the-art}
\newacronym{pac}{PAC}{Probably Approximately Correct}
\newacronym{asr}{ASR}{attack success rate}
\newacronym{iot}{IoT}{Internet of Things}

\newacronym{fp}{FP}{Factorized Prior}
\newacronym{mshp}{MSHP}{Mean Scale Hyper Prior}

\newacronym{bpp}{BPP}{bits per pixel}
\newacronym{dos}{DoS}{Denial of Service}
\newacronym{aml}{AML}{adversarial machine learning}
\usepackage{multirow}
\usepackage{algorithmic,algorithm}
\newcommand{\dnn}{\gls{dnn}\xspace}
\newcommand{\dnns}{\glspl{dnn}\xspace}

\usepackage{float}
\usepackage{wrapfig}

\usepackage[accsupp]{axessibility}  

\usepackage{hyperref}

\usepackage{orcidlink}

\begin{document}

\title{Resilience of Entropy Model in \newline Distributed Neural Networks} 

\titlerunning{Resilience of Entropy Model in Distributed Neural Networks}

\author{Milin Zhang \orcidlink{0009-0002-9675-8352} \and
Mohammad Abdi \orcidlink{0000-0001-6737-2632} \and  \newline
Shahriar Rifat \orcidlink{0000-0002-8111-2068} \and 
Francesco Restuccia\orcidlink{0000-0001-9757-7129} }

\authorrunning{M. Zhang et al.}

\institute{Institute for the Wireless Internet of Things \\ Northeastern University, Boston MA 02115, USA \\
\email{\{zhang.mil,abdi.mo,rifat.s,f.restuccia\}@northeastern.com}\\
}

\maketitle

\begin{abstract}
  Distributed \dnns have emerged as a key technique to reduce communication overhead without sacrificing performance in edge computing systems. Recently, entropy coding has been introduced to further reduce the communication overhead. The key idea is to train the distributed \dnn jointly with an entropy model, which is used as side information during inference time to adaptively encode latent representations into bit streams with variable length. To the best of our knowledge, the resilience of entropy models is yet to be investigated. As such, in this paper we formulate and investigate the resilience of entropy models to intentional interference (\eg, adversarial attacks) and unintentional interference (\eg,  weather changes and motion blur). Through an extensive experimental campaign with 3 different \dnn architectures, 2 entropy models and 4 rate-distortion trade-off factors, we demonstrate that the entropy attacks can increase the communication overhead by up to 95\%. By separating compression features in frequency and spatial domain, we propose a new defense mechanism that can reduce the transmission overhead of the attacked input by about 9\% compared to unperturbed data, with only about 2\% accuracy loss. Importantly, the proposed defense mechanism is a standalone approach which can be applied in conjunction with approaches such as adversarial training to further improve robustness. Code is available at \url{https://github.com/Restuccia-Group/EntropyR}. 
  \keywords{Trustworthy Machine Learning \and Data Compression}
\end{abstract}

\glsresetall

\section{Introduction}\label{sec:intro}

\begin{figure}[tb]
    \centering
    \includegraphics[width=.9\columnwidth]{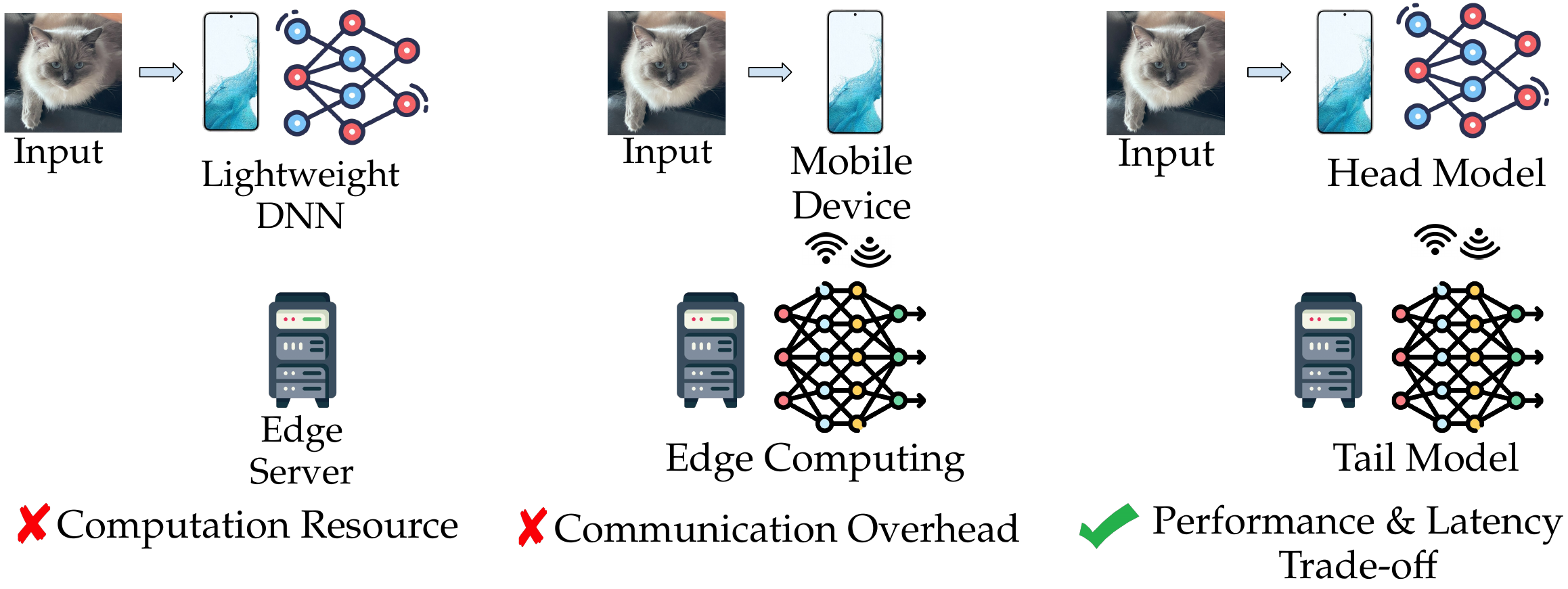}
    \caption{The key concept of distributed \dnn. Left: Lightweight \dnns suffer from performance loss due to the limited computational resources; Middle: Edge computing often results in intolerable latency due to communication overhead; Right: Distributed \dnns deploy a small \textit{head} model on the resource-limited device to achieve task-oriented compression and a large \textit{tail} model on the server to decode compressed features and execute the rest of model. \vspace{-0.5cm}}
    \label{fig:ddnn}
\end{figure}

Distributed \dnn were recently introduced to divide the computation of \dnns across various devices based on their available computation and communication resources. They have been shown to be extremely effective to implement deep learning applications on resource-constrained mobile devices~\cite{eshratifar2019bottlenet,shao2020bottlenet++,matsubara2022bottlefit}. As depicted in \cref{fig:ddnn}, the common strategy is to divide a large \dnn into a small \textit{head} network deployed on the mobile device to extract and compress features, and a \textit{tail} network on the server to perform the task inference. Compared to conventional lightweight \dnns specifically designed for mobile devices~\cite{howard2017mobilenets,sandler2018mobilenetv2,howard2019searching,tan2019mnasnet,jacob2018quantization}, distributed \dnns can leverage the computation power of the edge/cloud and hence attain better performance. In addition, distributed \dnns leverage compression techniques to reduce the data size of intermediate representations. By only transmitting compact representations, distributed \dnns can reduce the transmission overhead significantly compared to traditional edge computing~\cite{ran2018deepdecision,liu2018edgeeye,zhang2019masm}.

Interestingly, recent work has proposed to further minimize the data size of transmitted latent representations with entropy coding \cite{matsubara2022supervised}. The key idea is to train the distributed \dnn jointly with an auxiliary entropy model, which estimates the distribution of latent representations. The output of the entropy model is then used as side information during inference time to facilitate the adaptive encoding of latent representations into bit streams with variable lengths. The proposed approach achieves better compression rate compared to quantization-based~\cite{matsubara2020head} and codec-based~\cite{alvar2021pareto} approaches by minimizing the entropy of latent representations. \textit{Despite its excellent compression performance, the resilience of entropy coding to intentional interference (\eg, specifically crafted adversarial attacks) and unintentional interference (\eg,  sudden events such as weather changes and motion blur) remains unexplored}. We remark that in entropy-coding-based distributed \dnns, the bit rate of encoded representations relies completely on its estimated entropy. However, it is well known that \dnns trained on standard datasets are vulnerable to distribution shifts~\cite{hendrycks2018benchmarking,hendrycks2021many} and adversarial actions~\cite{szegedy2014intriguing,goodfellow2014explaining}. As the entropy model is learned without considering any corruptions and adversarial perturbations, \textit{a small change in the input space might lead to a large estimate of entropy, and hence result in a bit rate that can exceed the transmission bandwidth}. As depicted in \cref{tab:defense} in our experiments, the transmission overhead can be increased by about 2x in the worst case. 

In this work, we thoroughly assess for the first time the resilience of entropy models to intentional and unintentional interference as applied to distributed \dnns. Intriguingly, our findings unveil that the auxiliary entropy model learns a \textit{different set of input-related features} than what learned by the backbone \dnn. Ultimately, this enables us to design effective defense strategy to maintain the bit rate with an unnoticeable loss in the task performance. Our approach successfully reduces the communication overhead of perturbed data by about 9\% compared to unperturbed inputs, with only about 2\% accuracy loss. 

\subsection*{Summary of Novel Contributions}

$(1)$ \textit{We investigate for the first time the effect of intentional and unintentional interference to entropy coding in distributed \dnns.} We show that interference increases the end-to-end latency of distributed \dnns and poses a threat to other users by saturating the transmission bandwidth. Through comprehensive experiments involving 3 distinct \dnn architectures \cite{he2016deep,radosavovic2020designing}, 2 entropy models \cite{balle2017end,minnen2018joint}, and 4 rate-distortion trade-off factors, we illustrate that attacks to entropy coding may increase the transmitted data size by up to about 95\%; \smallskip

\noindent $(2)$ We design two visualization approaches to interpret the compression learned by entropy models. By disentangling features relevant to compression and classification across both frequency and spatial domains, we reveal that the entropy model learns specific features \textit{that are distinct} from those beneficial for the end task. This finding enables us to devise an effective defense strategy that safeguards the vulnerable compression features while minimally impacting task performance; \smallskip

\noindent $(3)$ We propose a defense approach to attacks targeting entropy coding based on our findings discussed above. The proposed method effectively reduces the transmission overhead of attacked inputs by about 9\%  in comparison to the unperturbed case, with only about 2\% decrease in accuracy. We remark that our approach is general in nature and can be combined with approaches such as adversarial training to further improve the resilience of entropy models. \smallskip

The paper is organized as follows. \cref{sec:background} provides background of entropy coding in distributed \dnns. In \cref{sec:model}, we formally describe the threat model for entropy coding in distributed \dnns. \cref{sec:exp} presents benchmarking results on both intentional and unintentional interference. \cref{sec:def} details the proposed defense approach and comprehensively evaluates it against adaptive attacks. Related work is summarized in \cref{sec:related_work}. Finally, we draw conclusions and discuss future work in \cref{sec:conclusion}. 

\section{Entropy Coding in Distributed DNNs}\label{sec:background}
\begin{figure}[tb]
    \centering
    \includegraphics[width=.95\columnwidth]{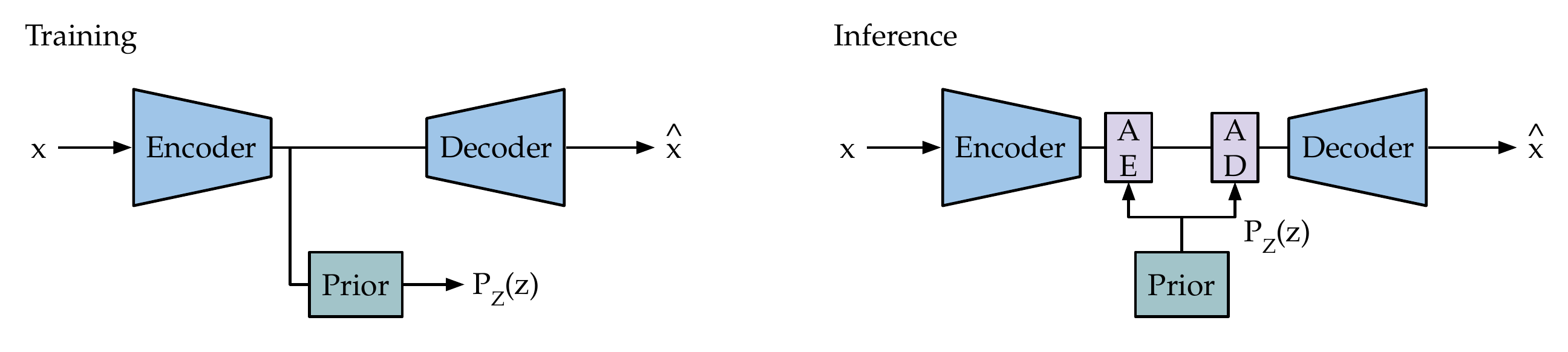}
    \caption{A general framework for entropy coding in neural data compression. It consists of an encoder-decoder structure to extract a compact latent representation $z$ and a prior model to estimate the distribution $P_Z(z)$. Left: During training, the prior model is jointly optimized with the encoder-decoder backbone. Right: During inference, the learned prior $P_Z(z)$ is used as side information for arithmetic encoding/decoding. }
    \label{fig:entropy_coding} 
\end{figure}

Entropy coding techniques such as arithmetic coding~\cite{rissanen1981universal} and asymmetric numeral systems~\cite{duda2013asymmetric} can encode a message with the optimal coding rate by leveraging the probabilistic information of the message. Based on the source coding theorem~\cite{shannon1948mathematical}, the entropy of a message $z$ with distribution $P_Z(z)$ defines the lower bound of expected coding rate $R_c(z)$ without any loss of information, i.e.,
\begin{equation}
    \mathbb{E}\{R_c(z)\} \geq H_{P_Z}(z) = \mathbb{E}\{-\log_2P_Z(z)\}
\end{equation}
where $H_{P_Z}(z)$ is the entropy of $z$. In recent data compression literature~\cite{balle2018variational,balle2017end,minnen2018joint,cheng2020learned,minnen2020channel,agustsson2020scale}, the entropy coding is integrated with lossy \gls{dnn}-based compression to attain an effective compression rate with less information loss compared to traditional approaches. As depicted in \cref{fig:entropy_coding}, the common strategy is to jointly train an encoder-decoder backbone for extracting compact latent representations $z$ with an auxiliary prior model $P_Z(z)$. During training, a random noise source is introduced to simulate the effect of quantization on $z$, allowing the backpropagation of gradients. In the inference phase, $z$ is initially quantized to integers for entropy coding. Next, the output of the prior model is used as side information to encode the quantized $z$ into bit streams adaptively with variable length. 

\gls{dnn}-based data compression can achieve the near-optimal compression rate by minimizing the entropy of latent representations during training. Specifically, the objective is to minimize the rate-distortion function
\begin{equation}
    \mathcal{L}(x,(\hat{x},z)) = \underbrace{||x-\hat{x}||^2_2}_\text{distortion} - \beta \cdot \underbrace{\log_2P_Z(z)}_\text{rate}, \label{eq:r-d}
\end{equation}
where $\hat{x}$ is the reconstructed data, $||x-\hat{x}||^2_2$ is the loss function of the encoder-decoder backbone and $-\log_2P_Z(z)$ is the loss function of the prior model. Constant $\beta$ denotes the trade-off between the rate and distortion. 

Recent work \cite{matsubara2022supervised} has incorporated entropy coding into distributed \dnns to increase communication efficiency between mobile and edge devices. The proposed method, referred to as the \textit{entropic student}, aims at achieving a balance between the coding rate and task-specific performance metrics (\eg, cross entropy in classification tasks). One distinction between the \textit{entropic student} and other entropy-coding-based feature compression work \cite{singh2020end,yuan2022feature} lies in the significantly smaller encoder size of the former due to constraints on computational resources in mobile devices. As a result, the \textit{entropic student} splits the \dnn at an early layer and use knowledge distillation~\cite{hinton2015distilling} for training the \textit{head} to preserve accuracy. \textbf{However, existing work does not address the issue of resilience of entropy models, which is the key goal of this paper}. A relevant work \cite{arvinte2023investigating} investigated the robustness of density estimation aiming to maximize $P_Z(z)$ while we are interested in perturbations that minimize $P_Z(z)$.

\section{Modeling Interference to Entropy Coding}\label{sec:model}

While the resilience of \dnns to distribution shifts \cite{hendrycks2018benchmarking,barbu2019objectnet,geirhos2018imagenet,wang2019learning,hendrycks2021many} and adversarial actions \cite{szegedy2014intriguing,carlini2017towards,madry2018towards,andriushchenko2020square,chen2020hopskipjumpattack,ilyas2018black} has been extensively investigated, we investigate types of interference that can alter the length of the encoded bit stream, which ultimately leads to increased bandwidth utilization. In this section, we formulate the resilience of compression in entropy-coding-driven distributed \dnns. 

\subsubsection{Threat Model for Accuracy.}
Let $D=\{x,y\}$ denotes a dataset for classification problem where $x \in \mathcal{X}$ and $y \in \mathcal{Y}$ are specific input and corresponding label samples respectively, while $\mathcal{X}$, $\mathcal{Y}$ are input and output space, respectively. There exists a prior yet unknown distribution $P(y|x)$ such that
\begin{equation}
    P(y|x) \geq P(\hat{y}|x) \qquad \forall\hat{y}\in \mathcal{Y}\label{eq:clf}
\end{equation}
where $\hat{y}$ denotes incorrect labels. The \dnn classifier, is trained to approximate the distribution $\Tilde{P}(y|x)\approx P(y|x)$ with empirical risk minimization approaches based on an assumption that in-sample data and out-of-sample data shared a similar distribution $P(y|x)$. However, the unintentional interference denoted by $\delta$, which can be modeled as an additive noise to $x$, creates a new dataset $D'=\{x+\delta,y\}$. It introduces a covariate shift that $P(x+\delta)\neq P(x)$ but $P(y|x+\delta)=P(y|x)$, resulting in classification errors such that
\begin{equation}
    \Tilde{P}(\hat{y}|x+\delta) \geq \Tilde{P}(y|x+\delta) \qquad \exists\hat{y}\in \mathcal{Y}\label{eq:interference}
\end{equation}


In the intentional interference scenario, algorithms attempt to find the minimum covariate shift that leads to \cref{eq:interference}. In this setting, a small perturbation in $l_p$ distance $||\delta||_{p} \leq \epsilon$ is intentionally crafted to mislead the \dnn classifier where $\epsilon$ denotes the constraint. Attackers formulate \cref{eq:interference} as an optimization problem over the input space $\mathcal{X}$, and hence can apply the gradient information of the loss function \wrt $x$ to find the $\delta$. \pgd \cite{madry2018towards} provided a unified view on iterative gradient attacks under $\mathit{l_{p}}$ constraint:
\begin{equation}
    x^{i+1} = x^i + \alpha\cdot\epsilon\cdot\frac{\triangledown_x\mathcal{L}}{||\triangledown_x\mathcal{L}||_p}\label{eq:pgd}
\end{equation}
where $x^{i}$ is the adversarial sample at $i$ step, $\alpha$ is the learning rate, and $\frac{\triangledown_x\mathcal{L}}{||\triangledown_x\mathcal{L}||_p}$ is the normalized gradient of a loss function $\mathcal{L}$ (\eg, a sign function in $l_\infty$ space). $\mathcal{L}$ can be cross entropy or other more advanced loss functions \cite{carlini2017towards}.

\subsubsection{Threat Model for Data Rate.} 

Let $H(x)=z$ denotes the \textit{head} network where $z\in \mathcal{Z}$ is the output of the \textit{head} and $\mathcal{Z}$ is the latent space. The goal is to achieve a minimum compression rate $-\log_2P_Z(z)$. As $z$ is dependent on $x$, the distribution shift introduced by unintentional interference can result in a difference in coding rate:
\begin{equation}
    -\log_2P_Z(H(x+\delta)) \neq -\log_2P_Z(H(x))\label{eq:compress}
\end{equation}
Note that the unintentional distribution shift of $x$ does not necessarily result in a larger bit rate. However, in the adversarial setting, attackers intend to \textit{maximize} the entropy of $z$ as follows:
\begin{equation}
\begin{aligned}
    \min_\delta \quad & \log_2P_Z(H(x+\delta)) \\
    s.t. \quad & ||\delta||_p \leq \epsilon
\end{aligned}
\end{equation}
Thus, to correctly assess the robustness of entropy models, we use the same \pgd algorithm in \cref{eq:pgd} yet with the entropy as its loss function $\mathcal{L} = -\log_2P_Z(H(x))$.

\section{Understanding the Resilience of Entropy Model}\label{sec:exp}

\subsection{Experimental Setup}

In this section, we outline the experimental setup for evaluating the resilience of entropy models. We first specify the \dnns and entropy models under consideration, followed by a description of the metrics and datasets used in our experiments. A summary of our experimental setup is provided in \cref{tab:exp}.

\begin{table}[tb]
  \caption{\dnns, entropy models and datasets utilized in experiments.}
  \label{tab:exp}
  \centering
  \begin{tabular}{@{}ll@{}}
    \toprule
    \multicolumn{2}{c}{\textbf{\dnn Architectures}}\\
    \midrule
    ResNet-50 \cite{he2016deep}& A standard baseline in distributed \dnns~\cite{matsubara2020head,matsubara2021neural,matsubara2022bottlefit,matsubara2022supervised} \\
    ResNet-101 \cite{he2016deep}& A deeper model in the ResNet family \\
    RegNetY-6.4GF \cite{radosavovic2020designing}& An advanced design with better performance \\
    \midrule    
    \multicolumn{2}{c}{\textbf{Entropy Models}}\\
    \midrule
    FP \cite{balle2017end} & A fully factorized prior model \\
    MSHP \cite{minnen2018joint} & An effective learnable entropy model \\
    \midrule
    \multicolumn{2}{c}{\textbf{Datasets}}\\
    \midrule
    ImageNet \cite{deng2009imagenet} & The standard ImageNet validation dataset \\
    ImageNet-C \cite{hendrycks2018benchmarking} & A synthetic dataset with 15 corruptions and 5 severities \\
    Random Noise & Noisy images as a baseline for adversarial robustness \\
    \pgd-Acc \cite{madry2018towards} & \pgd targeting classification \\
    \pgd-E & \pgd targeting entropy \\
  \bottomrule
  \end{tabular}
\end{table}

\subsubsection{Ablation of \dnns and Entropy Models.} 

We consider three distinct \dnns: ResNet-50~\cite{he2016deep}, ResNet-101~\cite{he2016deep} and RegNetY-6.4GF~\cite{radosavovic2020designing}. ResNet-50 serves as a common baseline in distributed \dnn literature~\cite{matsubara2020head,matsubara2021neural,matsubara2022bottlefit,matsubara2022supervised} while ResNet-101 is a deeper variant with the same design as former. RegNetY-6.4GF has an advanced design with better classification performance. The early layers are replaced by a specific tailored \textit{head} architecture as proposed in \cite{matsubara2022supervised} and incorporated with different entropy models. For the sake of brevity, we consider ResNet-50 as the default \dnn architecture without additional specification. We employ two entropy models: \gls{fp}  \cite{balle2017end} and \gls{mshp}  \cite{minnen2018joint}. \gls{fp} is the earliest design of entropy model and a basic component in many other advanced designs, while \gls{mshp} provides more powerful compression performance by injecting a hierarchical learnable block before \gls{fp} to extract the entropy information. For conciseness, we consider \gls{mshp} as the default entropy model. In addition, since models are trained with different rate-distortion trade-off $\beta$, we also investigate the compression robustness \wrt different $\beta$.

\subsubsection{Datasets and Metrics.}~Conversely from existing work that measures the compression performance with \gls{bpp}~\cite{balle2018variational,balle2017end,minnen2018joint,cheng2020learned,minnen2020channel}, we evaluate the entropy model with the data size of the whole bit stream after encoding as in \cite{matsubara2023sc2}. Compared to \gls{bpp},  data size quantifies more directly the networking traffic between the \textit{head} and \textit{tail}. To assess the resilience to unintentional interference, we use the ImageNet-C dataset proposed by~\cite{hendrycks2018benchmarking} comprising 15 common corruptions classified into 4 categories (noise, blur, digital, and weather) with 5 different severities. We also report the data size and accuracy on the clean ImageNet~\cite{deng2009imagenet} validation set as a baseline. To assess the resilience to intentional interference, we implement the \pgd targeting both the classification and compression performance in $l_\infty$ space. For clarity, we denote the conventional \pgd as \pgd-Acc and the \pgd targeting entropy models as \pgd-E. Meanwhile, random noise with the same perturbation level is added to the clean ImageNet as a baseline comparison for adversarial robustness.

\subsection{Resilience to Unintentional Interference}\label{sec:unintentional}

\begin{figure}[tb]
    \centering
    \begin{subfigure}{0.43\columnwidth}
    \includegraphics[width=\textwidth]{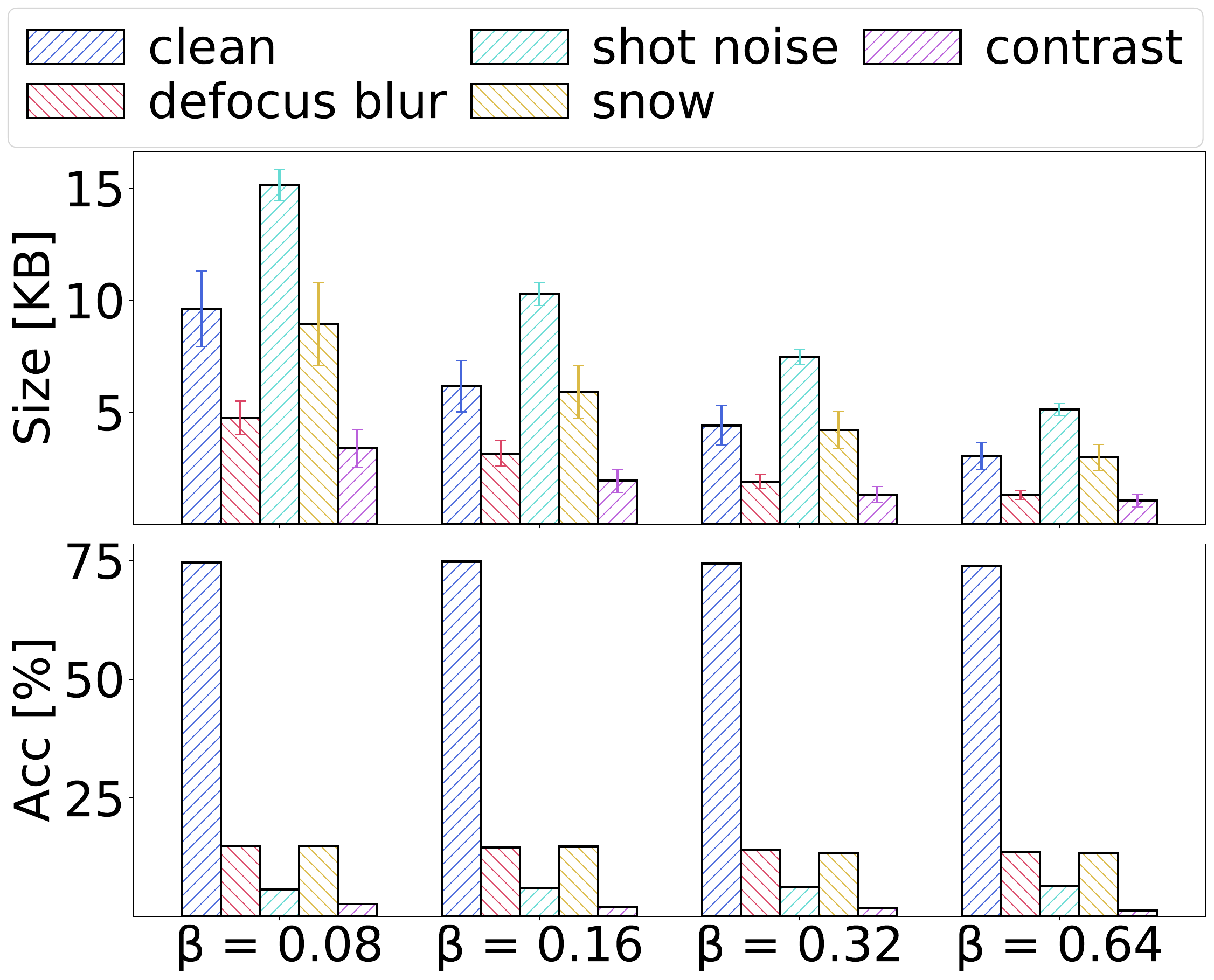}
    \caption{resilience \wrt $\beta$}
    \label{fig:c-beta}
    \end{subfigure}
    \begin{subfigure}{0.45\columnwidth}
    \includegraphics[width=\textwidth]{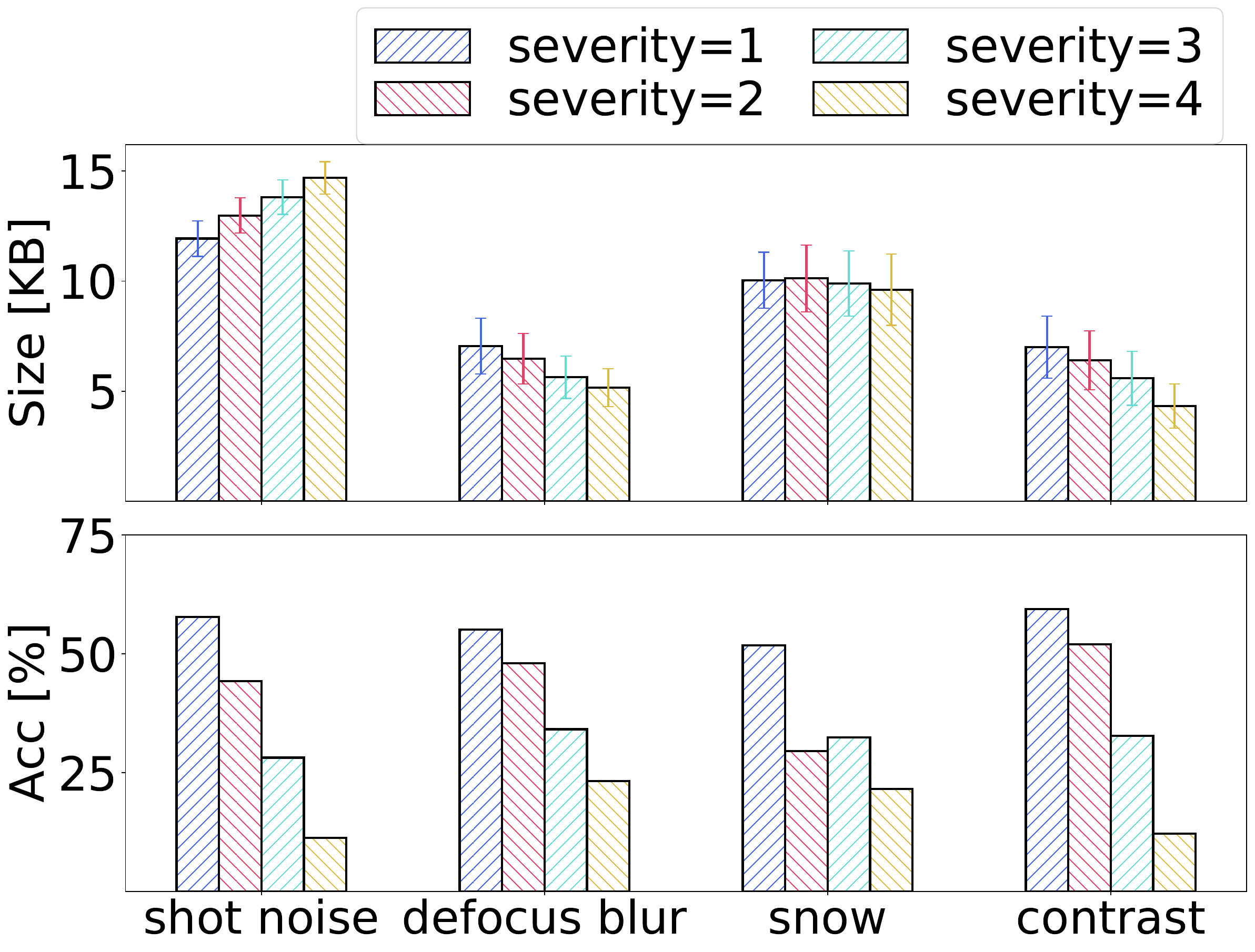}
    \caption{resilience \wrt severities}
    \label{fig:c-severity}
    \end{subfigure}
    \caption{Resilience of accuracy and data rate to 4 common corruptions.}
\end{figure}

\subsubsection{Resilience \wrt $\beta$.}

We select one corruption as a representative from each category in ImageNet-C~\cite{hendrycks2018benchmarking} to investigate the resilience as a function of rate-distortion trade-off $\beta$. \cref{fig:c-beta} shows the compression and classification performance on \textit{defocus blur}, \textit{shot noise}, \textit{snow} and \textit{contrast} dataset, each with the severity of 5. Compared to the clean ImageNet, all corruptions decrease the accuracy significantly. However, not all corruptions increase the data size. Only \textit{shot noise} increases the data size by 65.31\% on average. In contrast, \textit{Defocus blur} and \textit{contrast} decrease the data size by 53.31\% and 67.45\% on average, respectively. \textit{Snow} has little affect to compression, resulting in only 4.42\% decrease of data size. This indicates that \textit{the entropy model learns a different set of features than the classification model.} Since it shows the same trend for different rate-distortion trade-off factors, we only consider $\beta=0.08$ in next experiments. 

\subsubsection{Resilience \wrt Severities.}

Next, we explore resilience \wrt severity levels using the same corruptions mentioned earlier. As illustrated in \cref{fig:c-severity}, the data size of \textit{shot noise} rises with increasing severity, whereas \textit{defocus blur} and \textit{contrast} consistently decrease the data size. \textit{Snow} shows minimal variation across different severity levels. This can be attributed to \textit{defocus blur} and \textit{contrast} removing high-frequency components in images, while \textit{shot noise} introduces new patterns in high-frequency space. \textit{Snow}, on the other hand, lacks a specific pattern in the frequency domain. Thus, \textit{the compression is particularly sensitive to high-frequency information in the input space.}

\subsubsection{Resilience \wrt Prior Models.}

\cref{tab:corruption_prior} shows the impact of additional 4 corruptions at severity level 1 on both \gls{fp} and \gls{mshp} model. To further validate the previous observation, we choose 2 noise corruptions (\textit{Gaussian noise} and \textit{impulse noise}), which introduce high-frequency noise to images, and two blur corruptions (\textit{motion blur} and \textit{glass blur}), which eliminate high-frequency components from images. As shown in \cref{tab:corruption_prior}, \textit{Gaussian noise} increases 13.22\% and 24.32\% of data size for \gls{fp} and \gls{mshp}, respectively. \textit{Impulse noise} increases 14.42\% and 30.15\% of data size for \gls{fp} and \gls{mshp}, respectively. On the other hand, \textit{motion blur} reduces 9.79\% and 18.09\% of data size for \gls{fp} and \gls{mshp}, respectively. \textit{Glass blur} reduces 10.47\% and 19.13\% of data size for \gls{fp} and \gls{mshp}, respectively. Thus, the previous finding that compressive features reside in the high-frequency space remains consistent for different entropy models. 

\begin{table}[tb]
  \caption{Resilience \wrt prior models. The average data size of \gls{fp} and \gls{mshp} for clean ImageNet are $11.65\pm1.02$ and $9.62\pm1.70$ KBytes, respectively. \textcolor{red}{Red} for larger data size and \textcolor{blue}{blue} for lower compared to the baseline. }
  \label{tab:corruption_prior}
  \centering
  \begin{tabular}{@{}ccccccccc@{}}
    \toprule
    Prior & \multicolumn{2}{c}{Gaussian Noise} & \multicolumn{2}{c}{Motion Blur} & \multicolumn{2}{c}{Impulse Noise} & \multicolumn{2}{c}{Glass Blur} \\
    Model &Size[KB]&Acc[\%] &Size[KB]&Acc[\%] &Size[KB]&Acc[\%] &Size[KB]&Acc[\%] \\
    \midrule
    FP & \color{red}{13.19$\pm0.40$} & 59.51 & \color{blue}{10.51$\pm0.70$} & 62.22 & \color{red}{13.36$\pm0.34$} & 50.93 & \color{blue}{10.43$\pm0.66$} & 53.63 \\
    MSHP & \color{red}{11.96$\pm0.47$} & 59.36 & \color{blue}{7.88$\pm1.46$} & 61.58 & \color{red}{12.52$\pm0.35$} & 51.12 & \color{blue}{7.78$\pm1.39$} & 53.67 \\
  \bottomrule
  \end{tabular}
\end{table}

\subsection{Resilience to Intentional Interference}\label{sec:intentional}

\begin{figure}[tb]
    \centering
    \begin{subfigure}{0.45\columnwidth}
    \includegraphics[width=\textwidth]{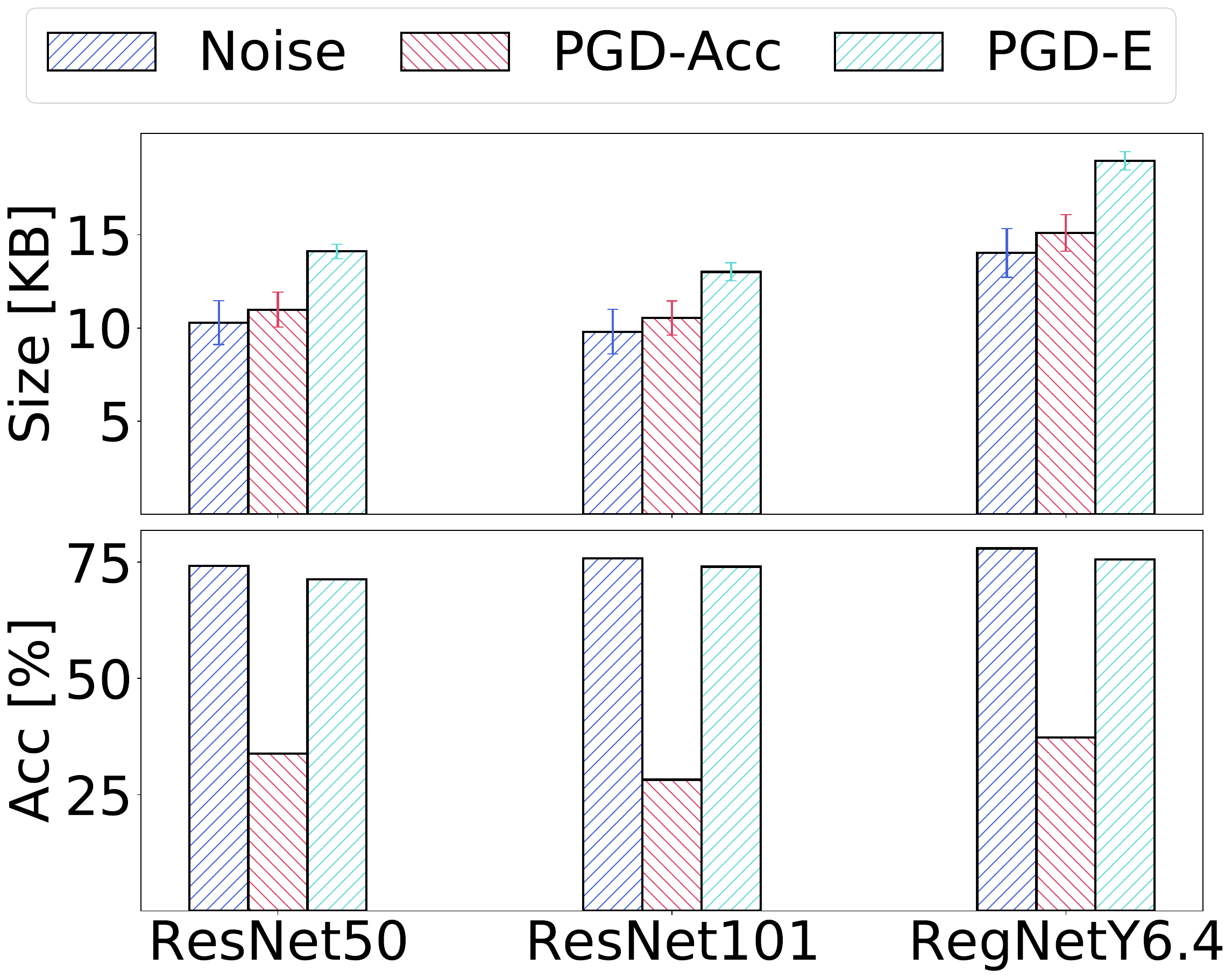}
    \caption{resilience \wrt \dnn architectures}
    \label{fig:atk-md}
    \end{subfigure}
    \begin{subfigure}{0.45\columnwidth}
    \includegraphics[width=\textwidth]{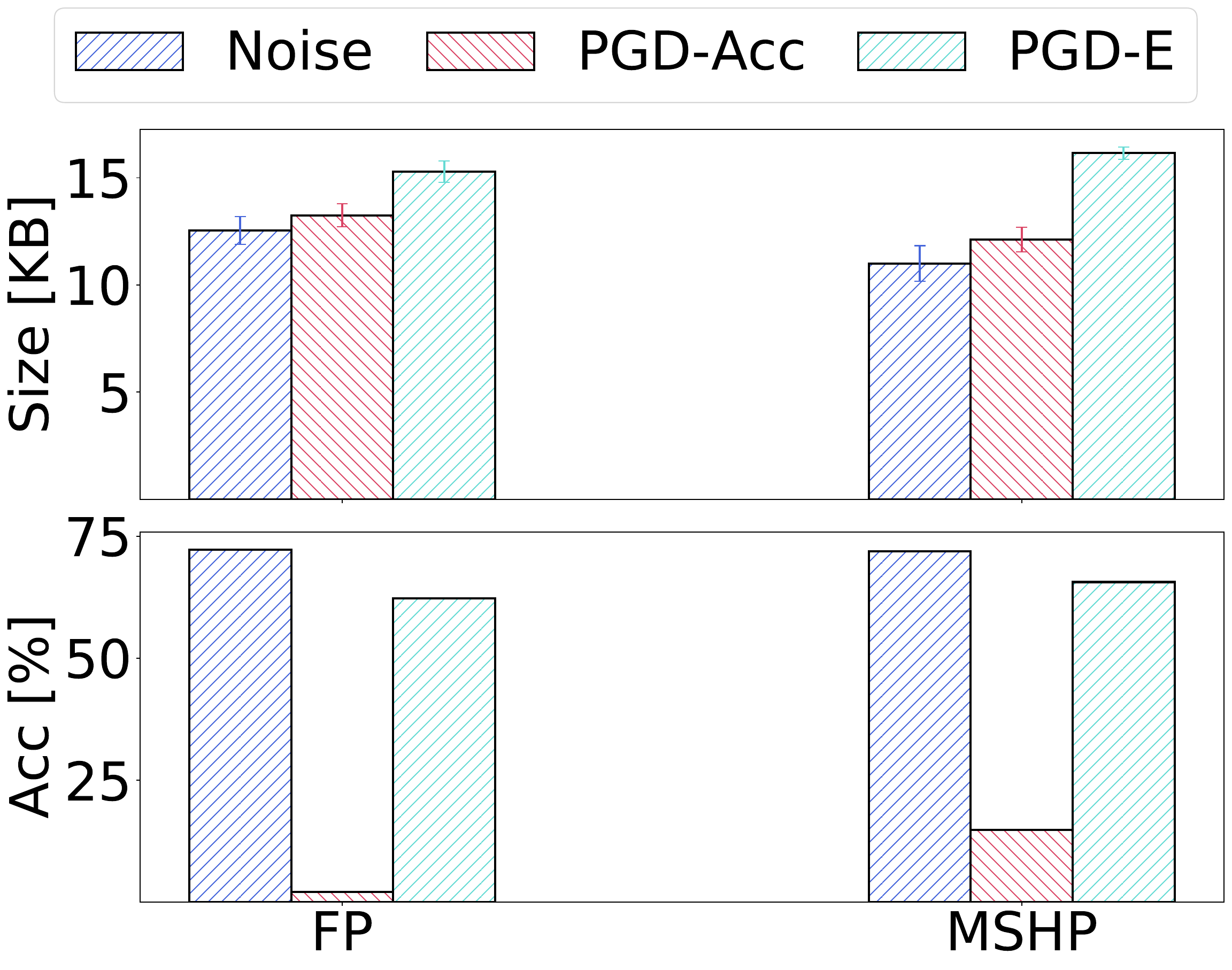}
    \caption{resilience \wrt prior models}
    \label{fig:atk-prior}
    \end{subfigure}
    \caption{Resilience of accuracy and data rate to intentional interference}
\end{figure}

\subsubsection{Resilience \wrt \dnns}

\cref{fig:atk-md} shows the resilience of different \dnn architectures to accuracy and entropy attacks with the $l_\infty$ constraint $\epsilon=4/255$. As evident in the figure, the attacks show similar patterns regardless of their victim models. 
While \pgd-Acc reduces 42.88\% accuracy on average, it only increases by 7.28\% of data size on average compared to random noise. On the other hand, \pgd-E decreases 2.34\% accuracy on average compared to random noise but increases the bit rate 1.35x times on average. This indicates that \textit{\pgd-E and \pgd-Acc are targeting separate features in the input space which has minor impact on each other.}

\subsubsection{Resilience \wrt Prior Models.}

\cref{fig:atk-prior} illustrates the effect of adversarial attacks with $\epsilon=8/255$ on victims using different prior models. As shown in the figure, both entropy models show considerable vulnerability to \pgd-E compared to \pgd-Acc. The data size are increased by 21.93\% and 46.82\% for \gls{fp} and \gls{mshp} respectively while the performance loss are 9.91\% and 6.62\% respectively. On the other hand, \pgd-Acc successfully reduces 72.10\% and 57.10\% classification performance of \gls{fp} and \gls{mshp} compared to random noise respectively. However, it only increases 5.66\% and 10.19\% bit rate for \gls{fp} and \gls{mshp} respectively. Hence, attackers targeting compression tend to perturb a different set of features than attackers targeting discriminative tasks regardless of the entropy models. In addition, while the \gls{fp} has a better robustness in compression, it has significantly worse robustness in accuracy.

\section{Proposed Defense Mechanism}\label{sec:def}

\subsection{Disentangle Compression Features in Frequency Domain}\label{sec:def1}

As shown in \cref{sec:unintentional}, interference that introduce high-frequency noise (\eg, \textit{shot noise}) can effectively increase the data size while interference that remove high-frequency information (\eg, \textit{defocus blur}) reduce the data size. Intuitively this is a consequence of the small \textit{head} in distributed \dnns as semantic information is usually captured in deeper layers while early layers tend to capture low level information \cite{bau2017network}.  To validate our intuition, we introduce total variation, defined as the integral of image gradient magnitude, which is first proposed for image denoising in~\cite{rudin1992nonlinear}. The anitrosopic total variation of a 2-D image is defined as
\begin{equation}
    TV(x) = \sum_{i,j}|x_{i+1,j} - x_{i,j}| + |x_{i,j+1} - x_{i,j}|
\end{equation}
where $x$ denotes an image sample, $i,j$ denotes the pixel position in width and height, respectively. The image gradient magnitude $|x_{i+1,j} - x_{i,j}|$ and $|x_{i,j+1} - x_{i,j}|$ describe the sharpness of pixel changes, where larger magnitude indicates high-frequency information such as edges and textures in images.

\begin{figure}[tb]
    \centering
    \includegraphics[width=0.9\columnwidth]{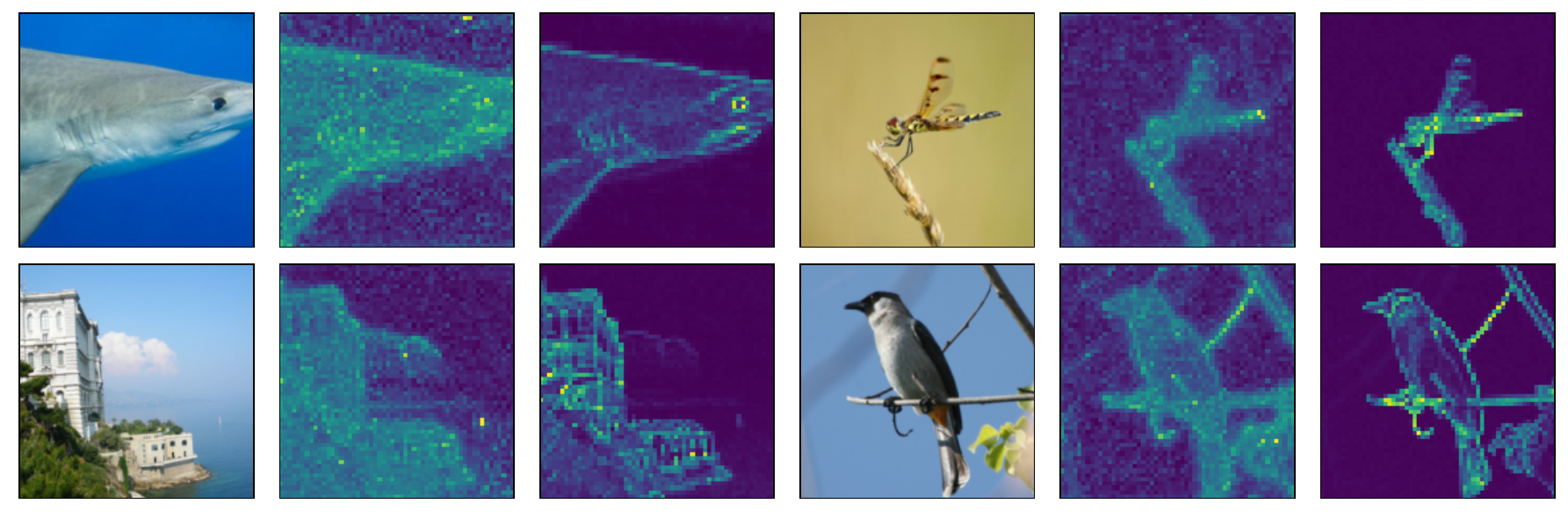}
    \caption{Entropy models are sensitive to high-frequency features. From left to right: images from ImageNet validation set, bit rate maps and total variation maps. }
    \label{fig:bpp_tv}
\end{figure}

We compare the total variation map of an image with its bit rate map generated by the entropy model. To plot the total variation map, we first slice an image into many small patches and then compute the total variation for each patch. Meanwhile, the bit rate map is the estimated entropy of the latent representation $-\log_2P_Z(z)$. As shown in \cref{fig:bpp_tv}, there exists a significant correlation between the bit rate map and the total variation map, which indicates that the entropy model is sensitive to high-frequency features in the input space.

\subsection{Disentangle Compression Features in Spatial Space}\label{sec:def2}

The work in \cite{matsubara2022supervised} has shown that entropy model in distributed \dnns tends to assign a larger number of bits to task-oriented information, such as objects, while compressing irrelevant information, like the background. Consequently, it becomes intuitively easier for an adversary to increase the bit rate in the background area to attack the entropy model.

\begin{figure}[tb]
    \centering
    \includegraphics[width=0.9\columnwidth]{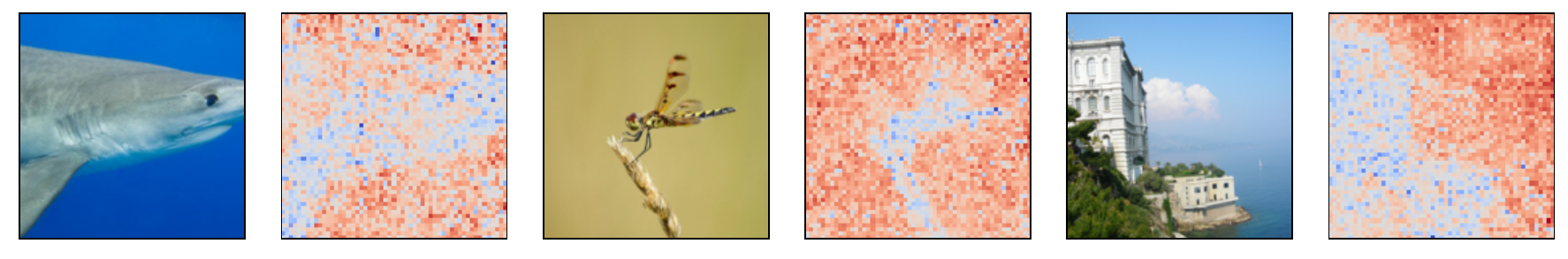}
    \caption{Entropy models are more vulnerable to perturbation in background region. From left to right: images from ImageNet validation set, bit rate comparison between \pgd-E and \pgd-Acc (\pgd-E allocate less bit in \textcolor{blue}{blue} area and more bit in \textcolor{red}{red} area). }
    \label{fig:bpp_atk}
\end{figure}

To interpret the action conducted by attackers, we generate the bit rate comparison map between adversarial samples generated by \pgd-E and \pgd-Acc.  This comparison map represents the estimated entropy of adversarial samples generated by \pgd-E subtracted from the bit rate of \pgd-Acc-perturbed samples. In this context, the red area indicates that \pgd-E tends to allocate more bits compared to \pgd-Acc, whereas the blue region suggests that \pgd-E tends to allocate fewer bits. As depicted in \cref{fig:bpp_atk}, \pgd-E puts more efforts to increase the bit rate in background region, whereas \pgd-Acc focuses on altering task-oriented information. This is in line with the experiments in \cref{sec:intentional} that \pgd-E and \pgd-Acc target different features in the input space. On one hand, increasing the bit rate in the background region will affect little accuracy as they are irrelevant information for the discriminative task. On the other hand, modifying object information can significantly degrade the performance, yet it has less impact on compression.

\subsection{Object-Aware Total Variation Denoising}

Given the observation in \cref{sec:def1}, we propose a denoising technique based on total variation to remove high-frequency noise in adversarial images where we solve the following optimization problem: 
\begin{equation}
    \min_x \frac{1}{2}||x - x'||^2_2 + \lambda\cdot TV(x),\label{eq:denoise}
\end{equation}
where $x'$ is the distorted image, $x$ is the denoised image, $TV(x)$ is the total variation of $x$ and $\lambda$ is a regularization factor to control the degree of smoothing. \cref{eq:denoise} means the image $x$ after denoising should keep most of the information in $x'$ (\ie, minimize $||x-x'||^2_2$) but discard the high-frequency information (\ie, minimize $TV(x)$). Optimizing \cref{eq:denoise} is not trivial as total variation is non-differentiable. As a result, we use a sub-gradient descent:
\begin{equation}
    x^{i+1} = x^{i} - \alpha \cdot ((x^i-x')+\lambda \cdot g(x^i))\label{eq:tvd}
\end{equation}
where $x^i$ is the denoised image at the step $i$, $\alpha$ is the step size and $g(x^i)$ is the image gradient of $x^i$ \cite{boigne2022towards}.

\textit{Simply applying the total variation denoising will also decrease the task-oriented performance as it also removes useful information in high-frequency domain}. Since adversarial attacks targeting entropy models attempt to increase bit rate in background region as demonstrated in \cref{sec:def2}, we add a mask to \cref{eq:tvd} to force the denoising algorithm to only remove high-frequency information in non-object space, that is,
\begin{equation}
    x^{i+1} = x^{i} - \alpha \cdot m\cdot((x^i-x')+\lambda \cdot g(x^i))\label{eq:tvd-r}
\end{equation}
where $m$ is the mask to control the denoising level. In practice, we interpolate the output of entropy model $P_Z(z)$ as the soft mask. This is because the higher bit rate $-\log_2P_Z(z)$ corresponding to object area has a smaller value of $P_Z(z)$, making it naturally a soft mask to avoid smoothing the object region. 

\subsubsection{Experimental Results.} We investigate the proposed defense method as a function of perturbation budget $\epsilon$. As shown in \cref{tab:defense}, attackers targeting entropy significantly increase the data size, by up to 95.32\% when $\epsilon=16/255$ compared to the clean images (9.62 KBytes). On the contrary, our denoising approach significantly reduce the data size for all $\epsilon$. For $\epsilon=2/255,4/255$, the average data size is even 8.94\% and 0.73\% smaller than the clean images with only 2.44\% and 1.18\% accuracy loss, respectively. For $\epsilon=8/255,16/255$, the data size is reduced by 31.80\% and 28.47\% respectively, while the accuracy improved slightly after denoising (1.60\% for $\epsilon=8/255$ and 7.74\% for $\epsilon=16/255$). 

We remark that the proposed defense is a standalone approach which can be incorporated into other approaches such as adversarial training to further improve the resilience. The accuracy loss incurred in small perturbation scenarios ($\epsilon=2/255,4/255$) can be mitigated by training with augmented data.

\begin{table}[tb]
    \centering
    \caption{Compression and classification performance before and after defense. The average data size for clean ImageNet is $9.62\pm1.70$ KBytes.}
    \begin{tabular}{@{}ccccccccc@{}}
    \toprule
     & \multicolumn{2}{c}{$\epsilon=2/255$} & \multicolumn{2}{c}{$\epsilon=4/255$} & \multicolumn{2}{c}{$\epsilon=8/255$} & \multicolumn{2}{c}{$\epsilon=16/255$} \\
     & Size[KB] & Acc[\%] & Size[KB] & Acc[\%] & Size[KB] & Acc[\%] & Size[KB] & Acc[\%] \\
    \midrule
    before & 12.35$\pm$0.63 & \textbf{73.26} & 14.11$\pm$0.38 & \textbf{71.33} & 16.15$\pm$0.29 & 65.62 & 18.79$\pm$0.37 & 52.13 \\
    after & \textbf{8.76$\pm$1.38} & 70.82 & \textbf{9.55$\pm$1.07} & 70.15 & \textbf{11.02$\pm$0.65} & \textbf{67.22} & \textbf{13.44$\pm$0.48} & \textbf{59.87} \\
    \bottomrule
    \end{tabular}
    \label{tab:defense}
\end{table}
\subsection{Adaptive Attacks}

To thoroughly evaluate the effectiveness of a defense strategy, \cite{carlini2017adversarial} proposed to design adaptive attacks that counter the defense mechanism with perfect knowledge of both \dnn and defense in the white-box setting. To this end, we assess our defense with two adaptive attacks. 

\subsubsection{Low Frequency Attack.}

As the total variation denoising is proposed to remove high-frequency components, the first adaptive attack that we considered is to only add perturbations in low frequency space. Following the methodology outlined in \cite{guo2018low}, we first convert the gradient to frequency space with discrete cosine transform and mask out the high-frequency components. The gradient is then transformed back and \cref{eq:pgd} is applied to create the adversarial samples.

\subsubsection{Regional Attack.} 

Given that the defense exclusively denoises the background region, the second adaptive attack we considered is to force the adversary put greater effort in increasing bit rate in the object region. Similar to \cref{eq:tvd-r}, a mask is multiplied to the loss function before backpropagation. In practice, we choose $1-P_Z(z)$ as the soft mask where $P_Z(z)$ is the probability of $z$ ranging between 0 and 1. A larger $P_Z(z)$ indicates a smaller bit rate $-\log_2P_Z(z)$ correponding to the background region. Therefore, $1-P_Z(z)$ encourages the adversary to take more attention to object region. 

\subsubsection{Resilience to Adaptive Attacks.} 
\cref{tab:adpt_atk} shows results of adaptive attacks with perturbation budget $\epsilon=4/255$. Our defense successfully reduces the average data size by 67.75\% and 67.62\% for low frequency attack and regional attack respectively while the accuracy only decreases 1.21\% and 1.41\% respectively. \textbf{Thus, the proposed approach is resilient to adaptive attacks. }

\begin{table}[tb]
    \caption{Resilience to adaptive attacks with $\epsilon=4/255$.}
    \centering
    \begin{tabular}{@{}ccccc@{}}
    \toprule
    Adaptive & \multicolumn{2}{c}{Before Denoising} & \multicolumn{2}{c}{After Denoising}\\
    Attacks & Size[KB] & Acc[\%] & Size[KB] & Acc[\%] \\
    \midrule
    Frequency & 14.11$\pm0.38$ & 71.38 & 9.56$\pm1.07$ & 70.17 \\
    Regional & 14.05$\pm0.41$ & 71.63 & 9.50$\pm1.10$ & 70.22 \\
    \bottomrule
    \end{tabular}
    \label{tab:adpt_atk}
\end{table}

\section{Related Work}\label{sec:related_work}

\subsubsection{Distributed Deep Neural Network.}

Distributed \dnn is proposed to meet the challenge of deploying artificial intelligence to resource-constrained platforms such as mobile devices and \gls{iot} devices~\cite{matsubara2022split}. To minimize the end-to-end latency across devices, dimension reduction designs (\ie, \textit{bottlenecks}) are introduced to compress the data size of intermediate representations that need to be sent \cite{eshratifar2019bottlenet,shao2020bottlenet++}. 

However, unlike autoencoders \cite{kingma2013auto,masci2011stacked,rifai2011contractive} that have symmetric designs on both encoder and decoder sides, distributed \dnns usually have asymmetric designs due to the limit of computation resources, and thus inject the \textit{bottlneck} in early layers~\cite{eshratifar2019bottlenet,shao2020bottlenet++,matsubara2022bottlefit,matsubara2020head,matsubara2021neural}. As a result, naively end-to-end trained distributed \dnns have noticeable performance loss \cite{eshratifar2019bottlenet,shao2020bottlenet++}. To preserve accuracy, \cite{matsubara2020head} proposed to use knowledge distillation to train the distributed \dnn separately while \cite{matsubara2022bottlefit} proposed a multi-stage training approach for each part of the \dnn. 

Along with \textit{bottlneck}-based distributed \dnns, different coding-based approaches are also applied to reduce the data size of the latent representations. \cite{alvar2021pareto} proposed to apply codec-based compression such as JPEG to latent representations and \cite{choi2018deep} adopted spatio-temporal coding such as HEVC to compress the streaming latent features in video tasks. \cite{singh2020end,yuan2022feature,matsubara2022supervised} use entropy coding to achieve a higher compression ratio of latent representations.

\subsubsection{Entropy Coding in Deep Data Compression.}

\cite{balle2017end} first introduced a fully factorized prior to integrate entropy coding with variational autoencoder, showcasing superior quality enhancement over conventional codec-based methods like JPEG for image compression. Building upon this, \cite{balle2018variational} extended the conventional factorized prior model to a learnable hierarchical hyper prior, resulting in improved performance. \cite{minnen2018joint} proposed a joint auto-regressive and hyper prior design while \cite{minnen2020channel} proposed a channel-aware entropy coding scheme. In addition, \cite{cheng2020learned} introduced the attention mechanism to prior models and \cite{agustsson2020scale} extended the technique to video compression. 

To minimize the communication overhead in distributed \dnn scenarios, \cite{singh2020end} for the first time adopted the entropy coding to further compress the latent representations after the \textit{bottleneck} while \cite{yuan2022feature} extended the approach to object detection. However, these approaches overlooked resource constraints, resulting in over-complex network designs for mobile devices. Conversely, \cite{matsubara2022supervised} integrated entropy coding with a multi-stage training strategy \cite{matsubara2022bottlefit}, aiming to optimize a tripartite trade-off encompassing computation resources in mobile devices, communication overhead between mobile and edge devices, and end-to-end performance of neural networks.

\subsubsection{Resilience of \dnn efficiency.}

While the reliability of \dnns have been extensively investigated, limited attention has been devoted to interference that undermine their efficiency. \cite{haque2020ilfo} explored this direction by demonstrating that imperceptible perturbations to input, leveraging intermediate representations of \dnns, could nullify the computation savings achieved by dynamic depth neural networks, which adapt their depth based on input complexity. \cite{pan2022gradauto} extended these attacks to target both dynamic depth and dynamic width neural networks. Moreover, in \cite{pan2023gradmdm}, the authors proposed simultaneous adjustments to the direction and magnitude of attacks, enhancing their effectiveness. \cite{hong2020panda} revealed that early exit dynamic \dnns could be tricked into late-stage inferences, significantly slowing down inference speed. 

The study of efficiency vulnerabilities in \dnns has also ventured into real-world deployment scenarios. \cite{liu2023slowlidar} focused on attacking LiDAR-based detection systems, introducing latency in detections and exposing vulnerabilities in critical contexts such as self-driving cars. \cite{chen2022nicgslowdown} demonstrated that neural image caption generation models could be manipulated to incur increased computation costs by inducing unnecessary decoder calls for token generation. 

In star contrast to existing research, our work investigates a novel threat emerging in distributed \dnns. Here, adversaries not only compromise the communication efficiency of entropy-coded distributed \dnns but also pose a threat to other users by saturating the transmission bandwidth.

\section{Conclusion}\label{sec:conclusion}

This paper has investigated the resilience of entropy models in distributed \dnns against both intentional and unintentional interference. We conducted thorough evaluations using 3 different \dnn architectures, 2 entropy model designs, and 4 rate-distortion trade-off factors with common corruption datasets and adversarial attacks. Our analysis disentangled compression features in both spatial and frequency domains, revealing vulnerabilities of the entropy model to specific types of perturbations. Building on these findings, we proposed a standalone defense strategy aimed at reducing data size with minimal task-oriented performance loss. Our future work will focus on designing more advanced defense approaches for distributed \dnns that are resilient in both compression and classification tasks.

\section*{Acknowledgements}

This work is funded in part by the Air Force Office of Scientific Research under contract number FA9550-23-1-0261, by the Office of Naval Research under contract number N00014-23-1-2221, as well as by National Science Foundation grants CNS-2134973 and CNS-2312875. Any opinions, findings and conclusions or recommendations expressed in this material are those of the authors and do not necessarily reflect the views of the U.S. Government.

%
%
\bibliographystyle{splncs04}
\bibliography{main}

\end{document}